\documentclass[twocolumn]{IEEEtran}

\usepackage{cite}

\ifCLASSINFOpdf

\else

\fi
\usepackage{graphicx}

\graphicspath{{images/}}
\usepackage{enumitem}
\usepackage{amsmath}
\usepackage{subfigure}
\usepackage{float}
\usepackage{cite}
\usepackage{svg}
\usepackage{amsmath}
\usepackage{graphics}
\usepackage{graphicx}
\usepackage{amssymb}
\usepackage{algorithm}

\usepackage{mathrsfs}
\usepackage{siunitx}
\usepackage{booktabs}
\usepackage{bm}
\usepackage{bbm}

\usepackage{float}
\usepackage{url}
\usepackage{amsfonts}
\usepackage[square,sort&compress,numbers]{natbib}
\usepackage{multirow}
\usepackage{textcomp}
\usepackage{supertabular}
\usepackage{array}
\usepackage{hyperref}
\def\degree{${}^{\circ}$}
\newcommand{\tabincell}[2]{\begin{tabular}{@{}#1@{}}#2\end{tabular}}

\usepackage{algorithm}  
\usepackage{algpseudocode}  
\usepackage{amsmath} 

\begin{document}
	
	\title{Geographical Knowledge-driven Representation Learning for Remote Sensing Images}

	\author{Wenyuan Li, Keyan Chen, Hao Chen and Zhenwei Shi$^\star$
		\IEEEmembership{Member,~IEEE}

		\thanks{Wenyuan Li,Keyan Chen, Hao Chen and Zhenwei Shi are with the Image Processing Center, School of Astronautics, Beihang University, Beijing 100191, China, and with the Beijing Key Laboratory of Digital Media, Beihang University, Beijing 100191, China, and also with the State Key Laboratory of Virtual Reality Technology and Systems, School of Astronautics, Beihang University, Beijing 100191, China.}
	}
	
	\maketitle
	
	\begin{abstract}
		The proliferation of remote sensing satellites has resulted in a massive amount of remote sensing images. However, due to human and material resource constraints, the vast majority of remote sensing images remain unlabeled. As a result, it cannot be applied to currently available deep learning methods.  To fully utilize the remaining unlabeled images, we propose a \textbf{Geo}graphical \textbf{K}nowledge-driven \textbf{R}epresentation learning method for remote sensing images (\textbf{GeoKR}), improving network performance and reduce the demand for annotated data.
		The global land cover products and geographical location associated with each remote sensing image are regarded as geographical knowledge to provide supervision for representation learning and network pre-training. An efficient pre-training framework is proposed to eliminate the supervision noises caused by imaging times and resolutions difference between remote sensing images and geographical knowledge. A large scale pre-training dataset Levir-KR is constructed to support network pre-training. It contains 1,431,950 remote sensing images from Gaofen series satellites with various resolutions.  Experimental results demonstrate that our proposed method outperforms ImageNet pre-training and self-supervised representation learning methods and significantly reduces the burden of data annotation on downstream tasks such as scene classification, semantic segmentation, object detection, and cloud / snow detection. It demonstrates that our proposed method can be used as a novel paradigm for pre-training neural networks. Codes will be available on \url{https://github.com/flyakon/Geographical-Knowledge-driven-Representaion-Learning}.
	\end{abstract}
	
	\begin{IEEEkeywords}
		representation learning, remote sensing images, scene classification, semantic segmentation, object detection, cloud / snow detection
	\end{IEEEkeywords}
	
	\IEEEpeerreviewmaketitle
	
	\section{Introduction}
	\label{sec:introduction}
	
	Due to the high capacity for learning features, deep learning methods have made significant progress in a variety of remote sensing image tasks, including object detection \cite{zou2017random,zhang2021salient,LI2018182,DING2018208}, cloud detection \cite{li2020matting,zou2019matting,LI2019197}, and semantic segmentation \cite{sun2018fully,yu2018semantic,ding2020semantic,KEMKER201860,DIAKOGIANNIS202094}. However, remote sensing images are increasingly exhibiting characteristics of multiple sources and resolutions. To achieve the best performance, images from each satellite must be annotated separately, which requires a significant amount of human and material resources. 
	
\begin{figure}[!htb]
		\centering
		\includegraphics[width=0.95\linewidth]{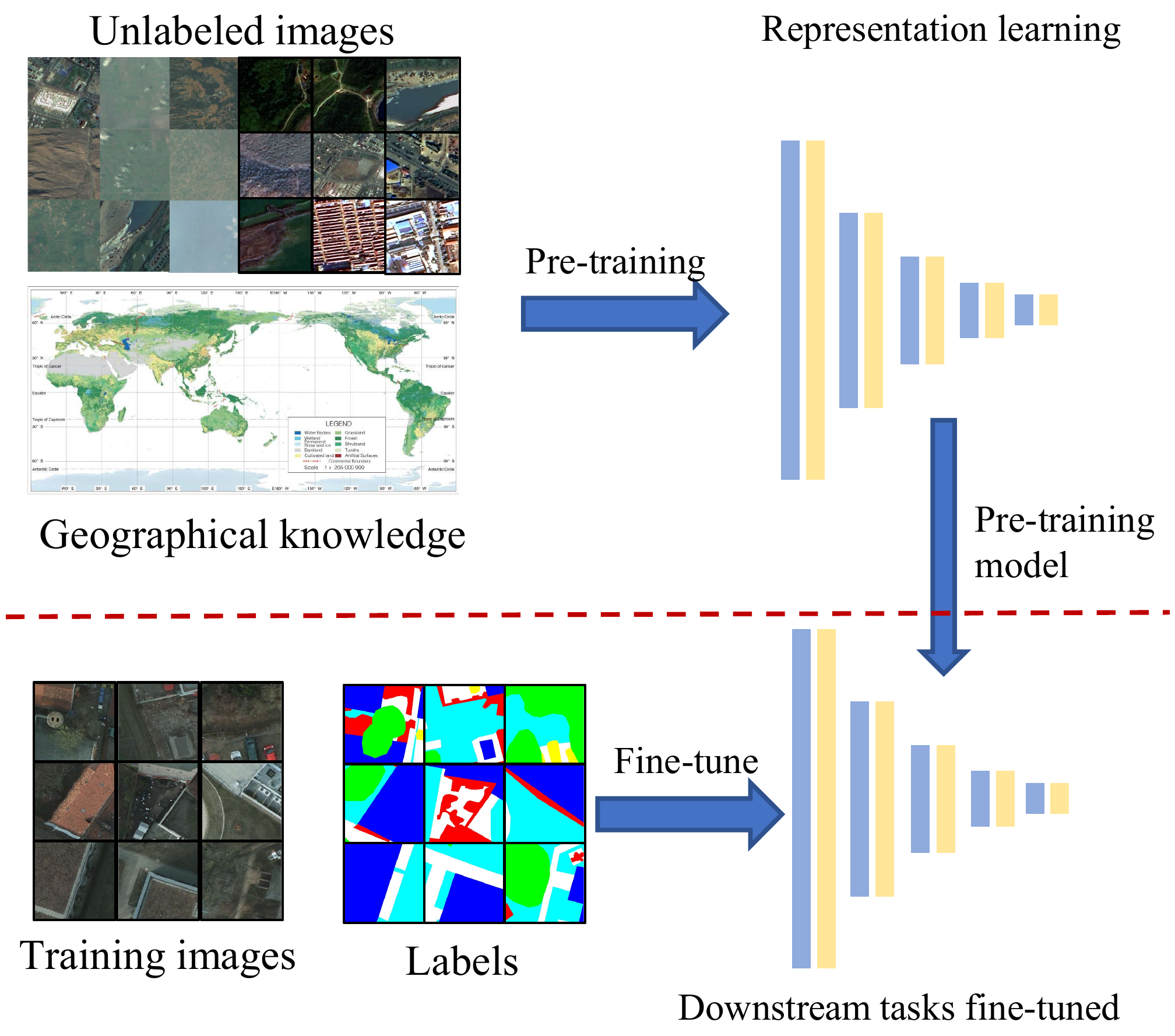}
		\caption{
			This is the subject we intend to study in this paper. A large number of unlabeled remote sensing images and geographical knowledge are utilized for representation learning and networks pre-training. With some annotated data, the pre-training model can be applied to downstream tasks. It has the potential to improve performance and alleviate the data annotation burden.
		}
		\label{fig:overall_method}
	\end{figure}

	Additionally, as the satellite circles the earth and monitors it, the same region will be photographed on a regular basis, generating new data on a continuous basis. But only a small portion of it will be annotated. Remaining unlabeled data contributes little to the improvements. The purpose of this paper is to study how to learn representations and pre-train networks using millions of unlabeled remote sensing images and pre-existing geographical knowledge. Geographical knowledge refers to the geographical location of remote sensing images and the global land cover product (GlobeLand30 \cite{jun2014open}). The pre-training models we obtain are then fine-tuned for a variety of downstream tasks (such as scene classification, semantic segmentation, object detection, and cloud detection) to improve performance and reduce the burden of data annotation. Fig. \ref{fig:overall_method} illustrates the overview of our proposed method \footnote{The figure of geographical knowledge cites from \url{https://www.webmap.cn/commres.do?method=globeDetails&type=brief}}.
	
	Nowadays, the most common approach for visual images is to use ImageNet \cite{deng2009imagenet} and JFT-300M \cite{Sun_2017_ICCV} datasets for pre-training and fine-tuning on downstream tasks. But both the ImageNet and JFT-300M datasets are manually annotated, and data annotation typically consumes a substantial amount of human and material resources. In order to make use of unlabeled data, self-supervised representation learning methods have been proposed and developed quickly. By the pretext tasks \cite{gidaris2018unsupervised,pathak2016context,jenni2018self,he2020momentum,chen2020simple,grill2020bootstrap}, it can help networks to learn image representations without the need for annotations.
	
	Self-supervised representation learning has also been used to improve the networks pre-training for remote sensing images \cite{vincenzi2020color,kang2020deep}. Recent researches \cite{ayush2020geography,manas2021seasonal} indicate that geo-location of remote sensing images may benefit self-supervised learning. These methods significantly enhance the effect by utilizing the unique auxiliary information contained in remote sensing images, which is extremely instructive. 
	
	However, the methods described above only make use of relatively simple geographical information, such as geographical location and imaging time. There is still a significant amount of geographical information available in the field of remote sensing.  We propose a \textbf{Geo}graphical \textbf{K}nowledge-driven \textbf{R}epresentation learning and network pre-training method for remote sensing images (\textbf{GeoKR}).  The GlobeLand30 and geographical location information associated with each remote sensing image are regarded as geographical knowledge to provide supervision for representation learning. GlobeLand30 is a global land cover product with a resolution of 30 meters that records the earth's ten major land covers. We can obtain land covers for each remote sensing image and map them into the knowledge representation using the geographical knowledge. The knowledge representation can be used as supervision and we can achieve efficient representation learning and network pre-training by aligning 
	image representations extracted from networks to the knowledge representation.  
	
	In addition, there are discrepancies between remote sensing images and geographical knowledge due to the difference in imaging times and resolutions, which may introduce random noises into the acquired supervision during the training phase.  Enlightened by the mean teacher method \cite{tarvainen2017mean} in noise labels and contrastive learning \cite{he2020momentum,chen2020improved}, we design an efficient representation learning framework.
	It consists of two networks with the exact same structure, dubbed the student and teacher networks. The teacher networks updates the weights from the student networks at a specified interval using the moving average method, which effectively reduces the drastic fluctuation of network parameters caused by noise supervision.
	
	We also create a pre-training dataset called Levir-KR. This dataset contains 1,431,950 images collected from the Gaofen-1, Gaofen-2, and Gaofen-6 satellites at various imaging resolutions and sources. The experimental section employs the teacher model as a pre-training model that is then fine-tuned for downstream tasks such as scene classification, semantic segmentation, object detection, and cloud / snow detection. Our method outperforms ImageNet pre-training, random initialization, and several recent self-supervised representation learning methods: MoCo \cite{he2020momentum}, SimCLR \cite{chen2020simple} and BYOL \cite{grill2020bootstrap}.  We also compared experimental results for each downstream task at various training data scales. Our experimental results demonstrate that our method performs significantly better with less training data, implying that it can effectively reduce the demand for data annotation for downstream tasks. Additionally, we conduct ablation analysis to determine the efficacy of the various components of our proposed method. All of the experiments demonstrate that our proposed method is an effective pre-training paradigm  for remote sensing images.
	
	The contributions are summarized as follows:
	
	\begin{itemize}
		\item We propose a \textbf{Geo}graphical \textbf{K}nowledge-driven \textbf{R}epresen-tation learning and networks pre-training method for remote sensing images (\textbf{GeoKR}). It can be considered a novel and effective method for pre-training networks, with the ability to improve the performance of downstream tasks
		\item  We developed a method for extracting supervision information from geographical knowledge and constructed an efficient representation learning framework for removing the influence of noise labels caused by discrepancies between remote sensing images and geographical knowledge.
		\item A pre-training dataset, Levir-KR,  is built. It contains 1,431,950 remote sensing images from Gaofen satellites with varying resolutions that can be used to effectively support network pre-training.
		\item The experimental results demonstrate that the proposed method outperforms random initialization, ImageNet pre-training, and other self-supervised representation learning methods in terms of improving downstream task performance and reducing the demand for annotated data.
	\end{itemize}
	
	The rest of this paper is organized as follows. In Section \ref{section:related work}, we introduce the related work. In Section \ref{section:method} and Section \ref{section:data_metric}, we give a detailed introduction of our proposed method and dataset. In Section \ref{section:experiment}, the experimental results are presented. Conclusions are drawn in section \ref{section:conclusion}.

	\section{Related Work}\label{section:related work}
	
	\subsection{Representation Learning}
	Representation Learning refers to learning representations of the data that make it easier to extract useful information when building classifiers or other predictors (also called downstream tasks) \cite{bengio2013representation}. Visual representation learning is the process of extracting visual representations from images or videos using deep learning methods (e.g., convolutional neural networks) to improve performance on image processing tasks. A good representation can be used as the input or as a pre-training model for supervised learning predictors. For instance, networks pre-trained on the ImageNet dataset \cite{deng2009imagenet} can develop a strong visual representation and are frequently used as a pre-training model for other image processing tasks, contributing significantly to the development of deep learning.
	
	Recently, the self-supervised representation learning method has made significant strides. Self-supervised representation learning methods \cite{goyal2019scaling,newell2020useful,kolesnikov2019revisiting} do not require manual annotations. They derive supervision information from the data itself via a series of pretext tasks. The design of pretext tasks is critical to the success of self-supervised learning. Initially, pretext tasks were constructed around the relationships between patches or objects in an image\cite{doersch2015unsupervised,santa2017deeppermnet,trinh2019selfie,kim2018learning,noroozi2016unsupervised,zhang2019aet,pathak2016context,larsson2016learning,larsson2017colorization,jenni2018self}.
	
	For example, occlude an area in the image randomly and repair it with networks, from which a good image representation is learned \cite{kim2018learning,pathak2016context,minderer2020automatic,jenni2018self}. \cite{larsson2017colorization,larsson2016learning} use image coloring as a pretext task directly. A pre-training model could be obtained while completing the task of image coloring. Jigsaw puzzles are another popular pretext task. It divides the image into patches and trains networks to understand it by predicting the relative position of each patch or by rearrangement of patches, from which image representations are learned \cite{doersch2015unsupervised,santa2017deeppermnet,noroozi2016unsupervised}. While the methods outlined above are effective at extracting visual representations from images, they are prone to overfitting.
	
	To increase the efficiency of representation learning, researchers are emphasizing high-level features of images when designing pretext tasks. Clustering and contrastive learning methods are now widely used. \cite{yang2016joint,xie2016unsupervised,kilinc2018learning,yan2020clusterfit,caron2018deep,huang2019unsupervised} generate pseudo labels using clustering methods and combine them with the loss function of networks during the training process, allowing networks to learn high-level representations.
	Contrastive learning \cite{chen2020simple,he2020momentum,chen2020improved} is the dominant method for self-supervised representation learning at the moment. It bases the pretext task on the following concepts: features between different images (typically referred to as negative samples) should be discriminative, but features between transformed data from the same image (typically referred to as positive samples) should be as similar as possible. Additionally, \cite{caron2020unsupervised} employs a combination of clustering and contrastive learning to achieve the highest accuracy on downstream tasks.
	
	\subsection{Representation Learning for Remote Sensing Images}
	Self-supervised representation learning researches are advancing rapidly for natural images at the moment, but are relatively scarce for remote sensing images. Vincenzi et al. \cite{vincenzi2020color} propose using high-dimensional data to reconstruct image color for pre-training, which they believe will aid networks in learning image representations. The SauMoCo \cite{kang2020deep} method successfully applies MoCo \cite{he2020momentum} to remote sensing images. 
	
	However, the above methods still adhere to natural image concepts, attempting to extract supervised information from remote sensing images through the use of pretext tasks similar to natural images, but do not take advantage of remote sensing images' characteristics. Recent researches have included geographical information in contrastive learning in order to achieve more efficient representation learning \cite{ayush2020geography,manas2021seasonal}. These methods have produced promising results and provided an excellent opportunity to investigate remote sensing representation learning. Inspire by these researches, we develop a novel representation learning paradigm for remote sensing images that is distinct from the one used for natural images.
	
	\subsection{Global Land Cover Products}
	
	Land cover information is critical for understanding climate change, ecological environment change, and etc. Recent advances in remote sensing technology and computer science have resulted in significant advances in mapping and research on global land cover (GLC) products with a spatial resolution of 30 meters \cite{liu2021finer}.  These products include GlobeLand30 \cite{jun2014open}, FROM\_GLC30 \cite{gong2013finer} and GLC\_FCS30 \cite{zhang2020glc_fcs30}. 
	
	The global land cover product identifies the different types of land cover that exist on the earth's land surface. GlobeLand30, for example, contains ten distinct land cover types: cultivated land, forest, grassland, shrubland, wetland, water body, tundra, artificial surface, bare land, and permanent snow, all of which have a spatial resolution of 30 meters and cover the world's major land areas. Each class is represented by a value among [10, 20, 30, 40, 50, 60, 70, 80, 90, 100]. Globeland30 consists of 849 TIF images spanning the globe's N85° - S85° latitude range. It uses the WGS-84 coordinate system. GlobeLand30's first version was released in 2000, and the most recent version is GlobeLand30 2020 \footnote{More details can be found at this website: \url{https://www.webmap.cn}}.
	
	Cao et al. \cite{cao2014preliminary}  analyze the global distribution of surface water resources and the spatial distribution and temporal fluctuations of surface water resources using data from GlobeLand30 2010 and GlobeLand30 2000. Chen et al. \cite{jun2015spatial} analyze the global urban and rural distributions using GlobeLand30, and used major indicators such as land area, composition ratio, and incremental ratio to quantify changes in urban and rural construction land in a variety of countries from 2000 to 2010. According to research, the United States and China accounted for roughly half of the global increase in new urban construction land over the last decade.
	
	Additionally, the global land cover products primarily reflect large-scale land cover types such as forest, grassland, and artificial surfaces. These land covers are relatively stable over relatively long periods of time. As a result, these products can be considered as auxiliary geographical knowledge for remote sensing representation learning.
	
	\begin{figure*}
		\centering
		\includegraphics[width=\linewidth]{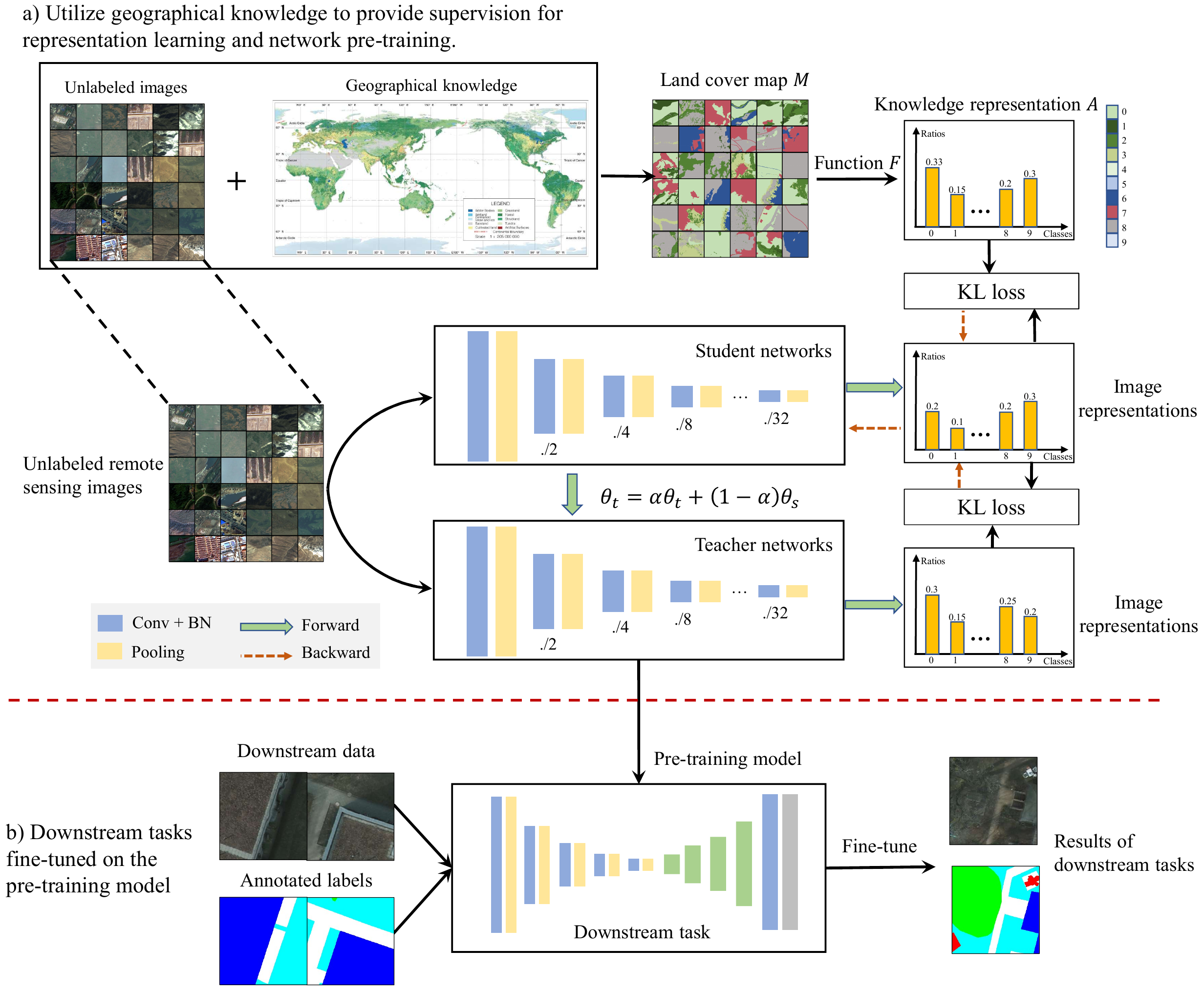}
		\caption{Details on our proposed method (GeoKR).
			a) the process of using geographical knowledge to provide supervision for representation learning and network pre-training.  b) process of fine-tuning on the pre-training model for downstream tasks.}
		\label{fig:overall_networks}
	\end{figure*}

	\section{The Proposed Method}\label{section:method}
	
	We propose a geographical knowledge-driven representation learning method for remote sensing images (GeoKR), which provides a novel paradigm for networks pre-training, as shown in Fig. \ref{fig:overall_networks}\footnote{The figure of geographical knowledge cites from \url{https://www.webmap.cn/commres.do?method=globeDetails&type=brief}}. The geographical knowledge we use is the global land cover product (GlobeLand30 \cite{jun2014open}) and geographical location of remote sensing images. We use geographical information to derive land cover types and proportions and then build a knowledge representation for each images. Aligning image representations extracted from networks with knowledge representations in representation space is the training object. The pre-training model can be fine-tuned with corresponding labeled data, improving downstream task performance.
	
	\subsection{Geographical Knowledge Supervision}
	
	GlobeLand30 and the geographical location with remote sensing images can be used to provide supervision for remote sensing representation learning. GlobeLand30 is a global land cover product that divides the earth into different areas and includes ten different land cover types.  Each area is saved as a separate tif file. GlobeLand30 does not contain any remote sensing image and can only be used to aid in the remote sensing representation learning when combined with other remote sensing images and their geographical location. Algorithm \ref{alg:main} illustrates the procedure for obtaining supervision information using geographical knowledge. 
	
	\begin{algorithm}  
		\caption{\label{alg:main}Supervision obtain process with geographical knowledge}  
		
		\begin{algorithmic}[1] 
			
			\State \textbf{Input:} remote sensing image $I$ of size $s$ and its geographical location $GT_I$, GlobeLand30 product $G$ and geographical location $GT_g^j$ of $j$th area $g_j$, the number of area $N$.
			\State \textbf{Output:} knowledge representation vector $A$ of image $I$.
			
			\Comment{firstly, determine to which area the image $I$ belongs.}
			\ForAll{$j \in \{1,2,\dotsc,N\}$}
			
			\State Determine whether $GT_I$ is in $GT_g^j$.
			
			\If {$GT_I  \in GT_g^j$}  
			\State $GT_g^m \gets GT_g^j$ 
			\EndIf
			\EndFor
			
			\Comment{secondly, calculate the relative coordinate of the image in the area.}
			
			\State calculate $x_{left}, x_{top}, x_{right}, x_{bottom}$ using Eq. \ref{equation:x_left} - Eq. \ref{equation:x_bottom}. 
			
			\Comment{finally, the proportion of different land covers is counted as the knowledge representation $A$}
			
			\State land cover map $M \gets g_j[x_{top}:x_{bottom},x_{left}:x_{right}]$
			
			\State calculate knowledge representation vector $A$
			\State \Return{$A$}
		\end{algorithmic}  
	\end{algorithm} 
	
	The first step is to find the area where the remote sensing image $I$ is located. It requires the remote sensing image's geographical location, as well as each area in GlobeLand30. A six-element vector $GT$ represents the geographical location information. $GT(0)$ and $GT(3)$ represent the image's upper left corner coordinates, while $GT(1)$ and $GT(5)$ represent the image's horizontal and vertical resolution alterations. The rotation resolution of the image is represented by $GT(2)$ and $GT(4)$, which are both 0. It is feasible to determine which area the remote sensing image is located in by finding $GT_g^m$ in the GlobeLand30 product that completely encircles the image area $GT_I$. 
	
	Then, the following formula can be used to determine the remote sensing image's relative position within the selected area $GT_g^m$:
	\begin{equation}\label{equation:x_left}
		x_{left} = \{GT_I(0)-GT_g^m(0)\}/{GT_g^m(1)},
	\end{equation}
	\begin{equation}\label{equation:x_top}
		x_{top} = \{GT_I(3)-GT_g^m(3)\}/{GT_g^m(5)},
	\end{equation}
	\begin{equation}\label{equation:x_right}
		x_{right} = x_{left}+ \{GT_I(1)*s\}/{GT_g^m(1)},
	\end{equation}
	\begin{equation}\label{equation:x_bottom}
		x_{bottom} = x_{top}+ \{GT_I(5)*s\}/{GT_g^m(5)},
	\end{equation}
	where $x_{left}, x_{top}, x_{right}, x_{bottom}$ represent the upper left and lower right coordinates of the image respectively. $s$ is the image size. The land cover map $M$ matching to the image $I$ can be generated from GlobeLand30 using the aforementioned location information. 
	
	$M$ records land cover types in remote sensing images. The following formula can be used to determine the quantity of each category:
	\begin{equation}
		S(i) = \sum_{x,y} {1_i(M(x,y)==i)}.
	\end{equation}
	$1_i$ represents pixel-wise indicator function and $i$ represents one category. If and only if the element in $M(x,y)$ is equal to $i$, the output of indicator function $1_i$ is $1$, otherwise it is $0$.  So $S(i)$ represents the quantity of category $i$ in land cover map $M$.

	We then design a transfer function $F$ to map $M$ into the knowledge representation $A$ as the supervision. The mechanism of mapping function $F$ is of great significance for the representation learning. Although there are numerous types of mapping functions $F$, the knowledge representation $A$ they generate must satisfy the following requirements: the similarity between knowledge representations is smaller for images with similar land covers, but larger for images with different land covers. We employ a straightforward transfer function: for each remote sensing image, we count the proportion of various land cover types as the knowledge representation $A$. Each element in $A$ can be calculated using the following formula:
	\begin{equation}
		A(i)=\frac{S(i)}{\sum_j S(j)}.
	\end{equation}
	
	It is obvious that the knowledge representation $A$ generated by this method satisfies the principles discussed above. 
	See section \ref{section:data_metric} for the rationality analysis of geographical knowledge supervision. 
	
	\subsection{Efficient Pre-training framework}
	There are discrepancies between remote sensing images and geographical knowledge due to the difference in imaging times and resolutions, which may introduce a certain amount of random noise in the acquired supervision during training phase. We build an effective pre-training framework based on mean-teacher networks in order to minimize the effects of supervision noises. It is made up of two networks with the same structure: the student networks and the teacher networks. 
	
	Encoder $f(\dotsc)$ and the projection head $g(\dotsc)$ are the two parts of the networks. The encoder $f$ extracts representations from remote sensing images. We adopt the widely used Resnet50 \cite{he2016deep} and VGG16 \cite{simonyan2014very} as the encoder to obtain $h_s=f_s(I), h_t=f_t(I)$, where $h_s$ and $h_t$ are the representations obtained by the student and teacher networks respectively. To project representations to the space where knowledge representation $A$ is located, we utilize a single fully connected layer as the projection head $g$ to obtain $S=g_s(h_s), T=g_t(h_t)$.
	
	The student networks are in charge of predicting and computing the loss function with supervision information, as well as updating the parameters using the gradient descent method. The teacher networks offer restrictions for the student networks and updates the parameters from the student networks at regular intervals using a moving average method.
	\begin{equation}
		\theta_t \gets \alpha \theta_t + (1-\alpha) \theta_s,
	\end{equation}
	where $\theta_t$ and $\theta_s$ represent the training weights of teacher networks and student networks respectively. $\alpha$ controls the update speed of the teacher network weights. 
	
	The KL (Kullback–Leibler) loss function is used to measure the similarity between image representation from networks and knowledge representation. 
	\begin{equation}
		L_{kr}=\sum_i A_i log\frac{A_i}{S_i},
	\end{equation}
	where $A$ represents the knowledge representation and $S$ represents the student networks outputs after operations of $softmax$. The above formula can be transformed into the following form:
	\begin{equation}
		L_{s}=-\sum_i A_i log S_i.
	\end{equation}
	
	We also use KL loss function to measure the distance between outputs of student networks and teacher networks:
	\begin{equation}
		L_{t}=-\sum_i S_i log T_i,
	\end{equation}
	where $S$ represents outputs of student networks and $T$ represents outputs of teacher networks. The total training function is as follows:
	\begin{equation}
		L=\gamma_1 L_{s} + \gamma_2 L_{t},
	\end{equation}
	where $\gamma_1$ and $\gamma_2$ are balance coefficients.
	As the teacher networks can avoid the fluctuation during the training, we use the teacher model as the pre-training model on downstream tasks.
	
	\subsection{Implementation Details}
	We develop our code using pytorch-1.5. The rate of learning is 1e-3. Following each epoch, the learning rate decreases to 90\% of the previous one. The batch size is configured to be 128. The teacher model updates the parameters once every 3000 batches, and the smoothing control parameter $\alpha$ is set to 0.95. The balance coefficients $\gamma_1$ and $\gamma_2$ are both set to 1.
	
	To increase the diversity of training data, we employ several data augmentation methods, including random rotation (rotation angle is chosen from [0\degree,90\degree,180\degree,270\degree), random flip up and down, and random flip left and right. Additionally, we use the color jitter method to increase the color diversity.
	
	\section{Dataset for Pre-training}\label{section:data_metric}
	
	\subsection{Introduction on Pre-training Dataset}
	We build up a pre-training dataset called Levir-KR.
	In this part, we introduce the statistical information of this dataset and the pre-processing methods to build it.  
	
	Levir-KR data are derived from the Gaofen series satellites, including Gaofen-1 fusion images with a resolution of 2 meters, Gaofen-1 multi-spectral images with a resolution of 16 meters, Gaofen-2 fusion images with a resolution of 0.8 meters, and Gaofen-6 multi-spectral images with a resolution of 16 meters. We convert the original scenes' front three bands to RGB (red, green, blue) mode. Then, with an overlap rate of 0.2, we cut them into images measuring $256 \times 256$.  Table \ref{tab:dataset_description} lists details about them. Columns from left to right represent the imaging sources, resolutions, number of original scenes and images. 
	
	\begin{table}
		\centering
		\caption{Details about the data we collect. Columns from left to right represent the data source, resolution, number of original scenes and images.}
		
		\begin{tabular}{c|c|c|c}
			\toprule
			\textbf{Data\ source} & \textbf{Resolution} &\textbf{Scenes} & \textbf{Images}\\
			\midrule
			Gaofen-1 fusion & 2 meter & 86 & 240994\\
			\midrule
			Gaofen-1 multi-spectral & 16 meter & 56 & 84365\\
			\midrule
			Gaofen-2  fusion & 0.8 meter & 25 & 1041502\\
			\midrule
			Gaofen-6 multi-spectral & 16 meter & 25 & 362835\\
			\midrule
		\end{tabular}
		\label{tab:dataset_description}
	\end{table}
	
	Due to the inescapable effect of clouds and sunlight on remote sensing images, we use unsupervised methods to remove images obscured by clouds or with low contrast. Given that the majority of clouds are brilliant white, we estimate the cloud proportion in each image using the following formula: 
	\begin{equation}
		r_c=\frac{1}{N}\sum\limits_i 1_t(R(i)>t),
	\end{equation}
	where $1_t(R(i)>t)$ indicates whether the pixel $i$ is covered by clouds. If the pixel value is greater than $t$, it is covered by clouds and we set $1_t=1$, otherwise it is not covered by clouds and we set $1_t=0$. The threshold $t$ is set to 230. $N$ represents the number of pixels. $r_c$ represents the proportion of clouds in the image. If the value is greater than 0.5, we discard this image.
	We use the method of scikit-learn \cite{scikit-learn} to judge whether the image is low contrast. If it is a low contrast image, we also directly discard it.
	
\begin{figure}
		\centering
		\includegraphics[width=0.75\linewidth]{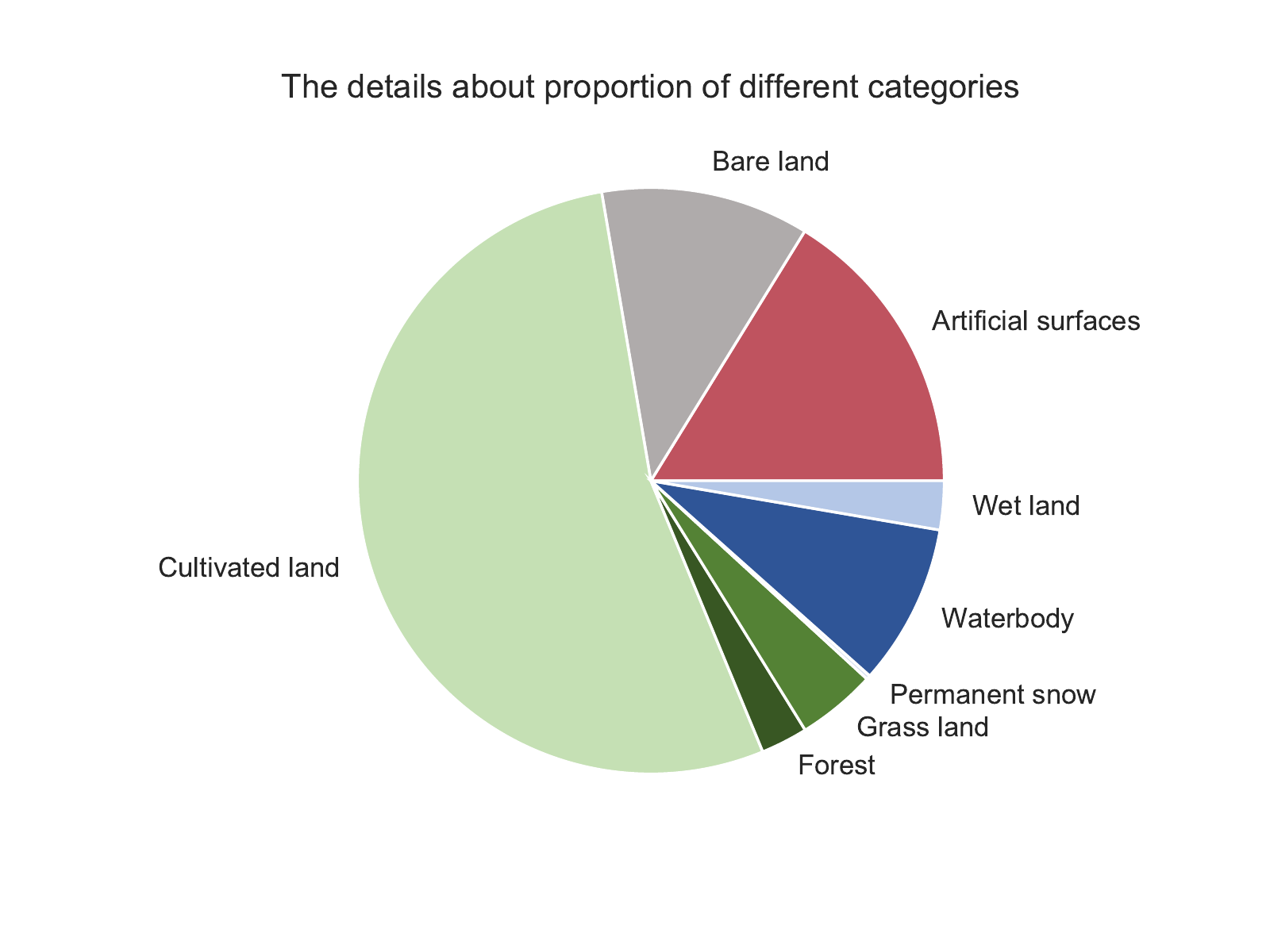}
		\caption{The details about proportion of different categories in Levir-KR dataset.  
		}
		\label{fig:dataset_overview}
	\end{figure}

	As it is difficult to collect enough remote sensing images of some rare categories, we discarded data of two categories: tundra and shrubland. Finally, we build up our pre-training dataset, Levir-KR, with 1,431,950 images.  Details about the Levir-KR dataset, including the name of each category, number of slices and proportion in all the data, are shown in the Table \ref{tab:labels_description} and Fig. \ref{fig:dataset_overview}. The category of each image is defined as one with the largest proportion.
	
	As a result of the uneven distribution of land cover on the actual land, the number of distinct classes is unbalanced. This will inevitably introduce bias into the pre-training model. As a result, during the network training process, we duplicate the images according to the number of different categories in order to balance the training data.
	
	\begin{table}
		\centering
		\caption{Details about categories of Levir-KR dataset. Columns from left to right are the name of each category, number of images and proportion in all the data.}
		
		\begin{tabular}{c|c|c|c}
			\toprule
			\textbf{Index} &\textbf{Class\ name} & \textbf{Number} &\textbf{Ratios}\\
			\midrule
			0 & Artificial \ surfaces & 232163 & 0.1621 \\
			\midrule
			1 & Bareland & 164365 & 0.1148 \\
			\midrule
			2 & Cultivated\ land & 766697 & 0.5354 \\
			\midrule
			3 & Forest & 37117 & 0.0259 \\
			\midrule
			4 & Grassland & 62419 & 0.0436 \\
			\midrule
			5 & Permanent\ snow & 3263 & 0.0023 \\
			\midrule
			6 & Water bodies & 127453 & 0.0890 \\
			\midrule
			7 & Wetland & 38473 & 0.0269 \\
			\midrule
		\end{tabular}
		\label{tab:labels_description}
	\end{table}
	
	\subsection{Analysis on Pre-training Dataset}
	
	The feasibility of using GlobeLand30 for pre-training will be discussed in this part. GlobeLand30 2000, GlobeLand30 2010, and GlobeLand30 2020 are the three versions of GlobeLand30. We use GlobeLand30 2010 and GlobeLand30 2020 to analyze changes in land covers over the last decade because the Levir-KR images are collected between 2010 and 2020. 
	
	\begin{figure}
		\centering
		\includegraphics[width=0.9\linewidth]{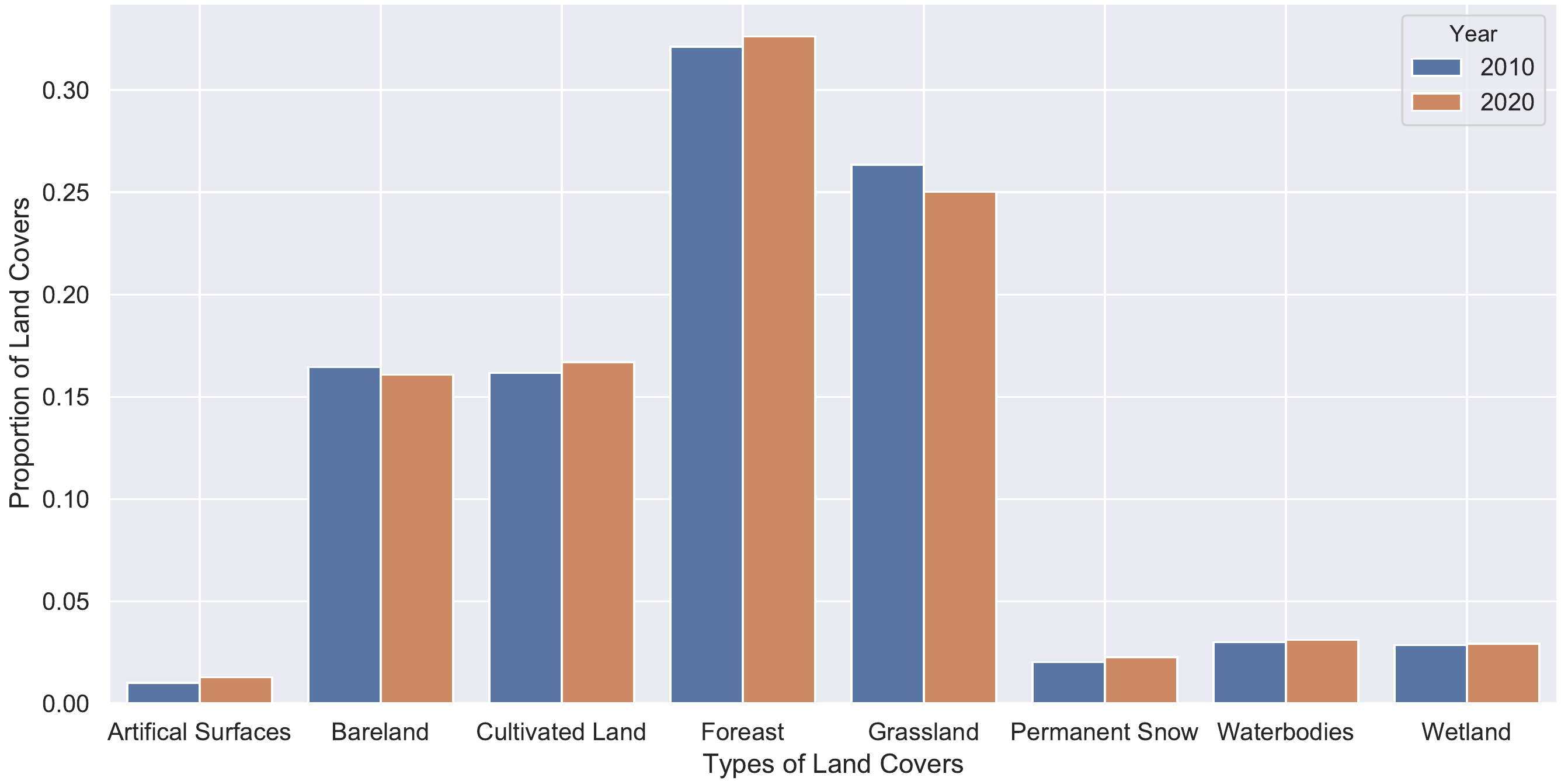}
		\caption{The proportion changes of land covers in GlobeLand30 2010 and GlobeLand30 2020.  
		}
		\label{fig:proportion_land_covers}
	\end{figure}
	\begin{table*}[!htb]
		\centering
		\caption{The proportions and changes of land covers generated by GlobeLand30 2010 and GlobeLand30 2020.}
		\resizebox{0.9\textwidth}{!}{
			\begin{tabular}{c|cccccccc}
				\toprule
				& \tabincell{c}{Artificial\\surfaces} & Bareland &\tabincell{c}{Cultivated\\land} & Forest & Grassland & \tabincell{c}{Permanent\\snow} &\tabincell{c}{Water\\bodies} &Wetland \\
				\midrule
				GlobeLand30 2010 & 0.0102 & 0.1645 & 0.1617 & 0.3212 & 0.2634 & 0.0204 &  0.0301 & 0.0285 \\
				GlobeLand30 2020 & 0.0129 & 0.1608 & 0.1669 & 0.3262 & 0.2501 & 0.0227 & 0.0311 & 0.0293 \\
				\midrule
				MAE & 0.0026 & 0.0037 & 0.0053 & 0.005 & 0.0133 & 0.0023 & 0.001 & 0.0007 \\
				MAPE & 0.2582 & 0.0227 & 0.0326 & 0.0156 & 0.0504 & 0.114 & 0.0341 & 0.0256 \\
				\midrule
		\end{tabular}}
		\label{tab:proportion_changes}
	\end{table*}
	
	We begin by calculating the changes in the eight land covers we used over the last decade, as shown in Fig. \ref{fig:proportion_land_covers} and Table \ref{tab:proportion_changes}. The table's first two rows depict the proportions of various land covers in the 2010 and 2020 versions. We evaluate land cover changes using "Mean Absolute Error (MAE)" and "Mean Absolute Percentage Error (MAPE)". Results are shown in the final two rows of Table \ref{tab:proportion_changes}. Apart from the relatively large changes in the artificial surfaces, the changes in other land covers over the last decade have been relatively small. Additionally, when combined with Fig. \ref{fig:proportion_land_covers}, despite the fact that the artificial surface changes are relatively large, their proportion in the dataset is small. Thus, our assumption that the majority of supervision information is accurate is confirmed, but there will be some noises. We can obtain a more intuitive understanding of this conclusion in Fig. \ref{fig:proportion_land_covers}. The histogram heights for various land covers are nearly identical in the 2010 and 2020 versions, indicating that the land covers have changed very little over the last decade.
	
	Additionally, we have analyzed changes in land covers in the Levir-KR dataset. Globeland30 2010 and 2020 are used to calculate the land covers of the images in the dataset, and the land cover with the highest proportion is chosen as the image category.  Then, we count the images with different categories in the two versions and calculate the percentage of images with category changes relative to the total number of images.  Our calculation indicates that the proportion is approximately 5\%, suggesting that the supervision information obtained using our proposed method is reasonably reliable.
	
\begin{figure}
		\centering
		\includegraphics[width=\linewidth]{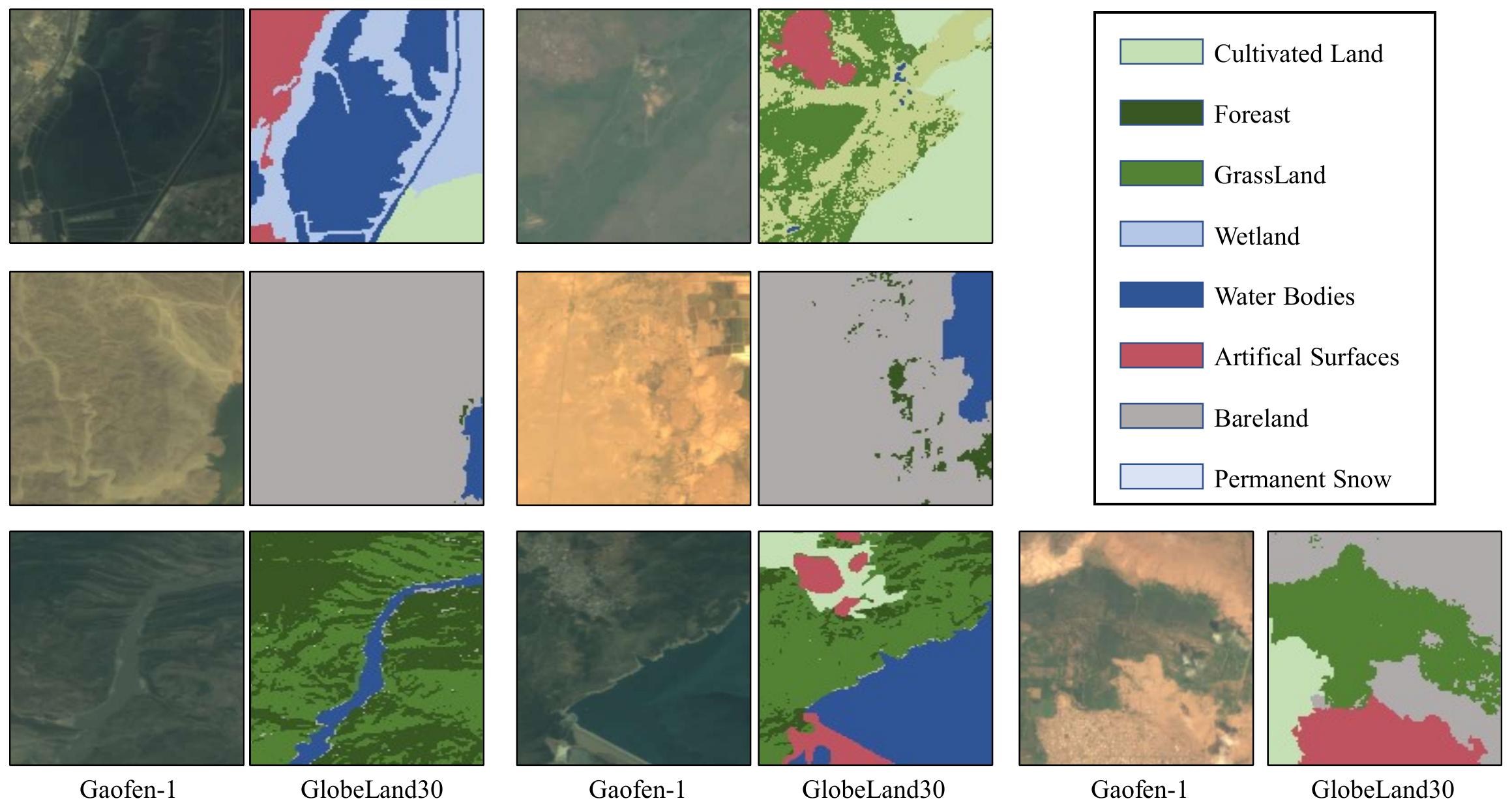}
		\caption{Some examples of GlobeLand30 2010 and corresponding Gaofen-1 images with the same location (Better viewed in color). The column "Gaofen-1" represents images from the Gaofen-1 satellite. The column "GlobeLand30" means the land cover map corresponding to the left Gaofen-1 images.}
		\label{fig:labelmap}
	\end{figure}

	Fig \ref{fig:labelmap} lists some examples in GlobeLand30 2010 and their corresponding Gaofen-1 images with the same location. Because of differences of imaging resolutions and imaging times between GlobeLand30 2010 and remote sensing images, the land covers maps can not be used for specific tasks. But they can also be roughly distinguished. Therefore, they can be used to provide a guide for remote sensing representation learning.
	
	\section{Experimental Results and Analysis}\label{section:experiment}
	In this section, our proposed method is evaluated on a variety of commonly used remote sensing image downstream tasks, including scene classification, semantic segmentation, object detection, and cloud / snow detection. Because our purpose is to verify the effect of representation learning on downstream tasks, we do not focus on designing the optimal network structure for downstream tasks, but rather on selecting a relatively simple network and comparing downstream task performance under various pre-training models. Random initialization, ImageNet pre-training and recently proposed self-supervised representation learning methods such as MoCo \cite{he2020momentum}, SimCLR \cite{chen2020simple}, BYOL \cite{grill2020bootstrap} are used as comparison pre-training methods in the experiment. All methods are trained on the Levir-KR dataset except for random initialization and ImageNet pre-training. 
	For each downstream task, we sample the downstream dataset separately to compare the fine-tuning effects of different pre-training models on different data scales. This is to ascertain the effectiveness of our method in terms of reducing the burden associated with data annotation. Additionally, our method includes two pre-training initialization methods: random initialization pre-training (represented by GeoKR) and continuing training based on the ImageNet pre-training model (represented by GeoKR$^\star$).  Only ResNet50 is used to evaluate the comparative self-supervised representation learning method. 
	
	\subsection{Scene Classification}
	
	\begin{table*}
		\caption{Scene classification results on UCMerced and RSSCN7. Best results are marked in bold. GeoKR$^\star$ indicates that it is continuously unsupervised  fine-tuned on the ImageNet pre-training model.}
		\resizebox{\linewidth}{!}{
			\begin{tabular}{c|ccccc|ccccc}
				\toprule
				\multicolumn{1}{c}{} &
				\multicolumn{5}{c}{\textbf{UCMerced}}  &  \multicolumn{5}{c}{\textbf{RSSCN7}}  \\
				& 5\% & 10\% & 20\% & 50\% & 100\% & 5\% & 10\% & 20\% & 50\% & 100\% \\
				\midrule
				VGG16 Random & 0.4552 &	0.5238 & 0.7752 & 0.8686 & 0.8705 & 0.5714 & 0.7586 & 0.8342 & 0.8814 & 0.8971 \\ 
				
				ResNet50 Random & 0.4247 & 0.5238 & 0.6362 & 0.7689 & 0.9105 & 0.4790 & 0.6143 & 0.6610 & 0.7179 & 0.8357 \\ 
				
				VGG16 ImageNet & \textbf{0.7238} & 0.8400 & 0.8781 & 0.8667 & 0.9448 & 0.7571 & 0.7729 & 0.8714 &	0.9057 & 0.9171 \\
				
				ResNet50 ImageNet & 0.6622 & 0.8273 & 0.8806 & 0.9022 & 0.9562 & 0.6752 & 0.7829 & 0.8276 & 0.8593 & 0.9229 \\
				\midrule
				
				MoCo \cite{he2020momentum} & 0.6533 & 0.7911 & 0.8381 & 0.9073 & 0.9378 & 0.6819 &  0.7790 & 0.8329 & 0.8710 & 0.9024 \\
				SimCLR \cite{chen2020simple} & 0.5746 & 0.7403 & 0.8127 & 0.9168 & 0.9365 & 0.6914 & 0.7581 & 0.7919 & 0.8138 & 0.8710 \\
				
				BYOL \cite{grill2020bootstrap} & 0.5702 & 0.7632 & 0.8787 & 0.8737 & 0.9422 &  0.6948 & 0.7748 & 0.7733 & 0.8252 & 0.8714 \\
				\midrule
				GeoKR (VGG16) & 0.6286 & 0.7905 & 0.8610 & 0.9346 & 0.9575 & 0.8333 & 0.8600 & 0.8771 & 0.9204 & \textbf{0.9443} \\
				GeoKR (ResNet50)  & 0.6229 & 0.7867 & 0.8368 & 0.8978 & 0.9416 & 0.8029 & 0.8548 & 0.8843 & 0.8924 & 0.9005\\
				
				GeoKR$^\star$ (VGG16) & 0.6330 & 0.8197 & 0.8673 & 0.9276 & \textbf{0.9702} & \textbf{0.8481} & 0.8662 & 0.8748 & \textbf{0.9210} & \textbf{0.9443} \\
				
				GeoKR$^\star$ (ResNet50) & 0.7048 & \textbf{0.8470} & \textbf{0.9092} & \textbf{0.9549} & 0.9695 & 0.8448 & \textbf{0.8933} & \textbf{0.8933} & 0.9152 & 0.9419 \\
				\bottomrule
				
		\end{tabular}}
		
		\label{tab:scence_classification}
	\end{table*}
	
	\textbf{Experimental setup}. The UCMerced \cite{yang2010bag} and RSSCN7 \cite{zou2015deep} datasets are used to evaluate our proposed model for scene classification task. The UCMerced dataset contains 21 categories with 100 images each, while the RSSCN7 dataset contains seven categories with 400 images each. The training and testing sets were randomly divided in a 3:1 ratio. All weights have been fine-tuned, including the backbone and linear classification layer. Each of the comparative methods utilizes the identical network structure and training strategy. The rate of learning is set to 0.001, and a total of 200 epochs are trained. Random flip and random rotation are added to increase the diversity of the data. The mean average of the top-1 accuracy of each class is used as evaluation metric. Additionally, we repeated each experiment three times and took the average of the three results as the final results.
	
	\textbf{Experimental results}. The experimental results are shown in the Table \ref{tab:scence_classification}. The best results are marked in bold. Each column denotes the results obtained using various proportions of training data. Our method achieves the best results in almost all cases, as demonstrated by the experimental results. 
	
	As the most frequently used method, ImageNet pre-training is still superior to other comparison methods. When all training data is used, however, our method outperforms ImageNet pre-training by two percentage, demonstrating that our method can indeed improve scene classification performance. Self-supervised learning methods fall between "ResNet50 Random" and "ResNet50 ImageNet" in terms of performance. This demonstrates that while the current self-supervised representation learning method can improve scene classification performance, adding geographical knowledge can improve performance even further. 
	
	In addition, when using only 50\% labeled data for scene classification, our method can obtain the comparable performance as ImageNet pre-training method with 100\% labeled data. It demonstrates that our method can alleviate some of the burden associated with data annotation. In practice, only half of the original work may be required to achieve the desired performance.
	
	\subsection{Semantic Segmentation}
	
	\textbf{Experimental setup}. We demonstrate our method's effectiveness in semantic segmentation of remote sensing images using the Potsdam and Vaihingen datasets \cite{rottensteiner2012isprs}. Six categories are included in the datasets. Each dataset is randomly divided into a training set, a validation set, and a testing set in the ratio 3:1:1. Each image in the datasets is cut into slices with $256 \times 256$ size. We construct simple fully convolutional neural networks \cite{long2015fully} and incrementally up-sample feature maps using the bilinear layer. Each bilinear layer is followed by a convolutional layer and a batch normalization layer \cite{ioffe2015batch}, which doubles the size of the output feature maps. During training, the learning rate is set to 0.005 and decreases to 90\% of the previous value every ten epochs. The model is trained over a period of 200 epochs. To increase generalization ability, random flip and random rotation are added. As an evaluation metric for the semantic segmentation task, we use mIoU (mean average of Intersection-over-Union). Meanwhile, we will calculate the accuracy of the validation set every 20 epochs and choose the model with the highest accuracy as the final one.   
	
	\textbf{Experimental results}. The segmentation results are shown in the Table \ref{tab:vaihingen_segmentation_iou} and Table \ref{tab:potsdam_segmentation_iou}. The best results are marked in bold. Each column represents the results with different proportions of training data. Experimental results demonstrate that our method consistently produces the best results, and the improvement is more pronounced with less training data.  
	
	In comparison to random initialization, ImageNet pre-training does not significantly improve semantic segmentation accuracy, whereas self-supervised representation learning methods can in some cases outperform ImageNet pre-training. The performance of different comparison methods is approaching saturation as the amount of training data grows, but our method still improves by about 4\% on the Vaihingen dataset and 2\% on the Potsdam dataset. It demonstrates the critical role of geographical knowledge in representation learning for remote sensing semantic segmentation.
	
	In addition, when using only 20\% labeled data for semantic segmentation, our method can obtain the comparable performance as ImageNet pre-training method with 100\% labeled data. 
	
	\begin{table*}
		\centering
		\caption{Sementaic segmentation results on Vaihingen dataset. mIoU is used as the evaluation index. Best results are marked in bold. GeoKR$^\star$ indicates that it is continuously unsupervised  fine-tuned on the  ImageNet pre-training model.}
		\resizebox{\textwidth}{!}{
			\begin{tabular}{c|cccccccccc}
				\toprule
				& 0.25\% & 0.33\% & 0.5\% & 1\% & 2\% & 5\% & 10\% & 20\% & 50\% & 100\% \\
				\midrule
				VGG16 Random & 0.3600 & 0.3207 & 0.3540 & 0.3810 & 0.4041 & 0.4596 & 0.4770 & 0.5024 & 0.6313 & 0.6521 \\ 
				
				ResNet50 Random & 0.3054 & 0.3439 & 0.3369 & 0.3846 & 0.3757 & 0.4194 & 0.4727 & 0.5106 & 0.6309 & 0.6448 \\ 
				
				VGG16 ImageNet & 0.3218 & 0.3430 & 0.3777 & 0.3866 & 0.3855 & 0.4731 & 0.4807 & 0.5028 & 0.6293 & 0.6753 \\
				
				ResNet50 ImageNet & 0.2974 & 0.3424 & 0.3575 & 0.3470 & 0.4050 & 0.4640 & 0.4455 & 0.5177 & 0.6611 & 0.7015 \\
				\midrule
				
				MoCo \cite{he2020momentum} & 0.3295 & 0.3407 & 0.3463 & 0.4800 & 0.4354 & 0.5450 & 0.5929  & 0.6128 & 0.6406 & 0.6819\\
				
				SimCLR \cite{chen2020simple} & 0.3270 & 0.2609 & 0.3349 & 0.4102 & 0.4450 &  0.5098 & 0.5939 & 0.5979 & 0.6417 & 0.6651\\
				
				BYOL \cite{grill2020bootstrap} & 0.2325 & 0.2694 & 0.3179 & 0.3976 & 0.4913 & 0.5925 & 0.6447 & 0.6829 & 0.6870 & 0.7271 \\
				\midrule
				GeoKR (VGG16) & 0.3480 & 0.3530 & 0.4244 & 0.4826 & 0.5423 & 0.6023 & 0.6347 & 0.6749 & 0.6999 & 0.7210 \\
				GeoKR (ResNet50)  & \textbf{0.3634} & 0.3152 & \textbf{0.4285} & \textbf{0.5165} & \textbf{0.5783} & 0.6209 & 0.6423 & 0.6796 & 0.6861 & 0.7110\\
				GeoKR$^\star$ (VGG16) & 0.3405 & 0.3738 & 0.4176 & 0.4812 & 0.5491 & 0.6185 & \textbf{0.6704} & \textbf{0.6978} & 0.7154 & \textbf{0.7401} \\
				
				GeoKR$^\star$  (ResNet50) & 0.3607 & \textbf{0.4138} & 0.4155 & 0.5150 & 0.5665 & \textbf{0.6390} & 0.6397 & \textbf{0.6978} & \textbf{0.7159} & 0.7268 \\
				\bottomrule
				
		\end{tabular}}
		
		\label{tab:vaihingen_segmentation_iou}
	\end{table*}
	
	\begin{table*}
		\centering
		\caption{Sementaic segmentation results on Potsdam dataset. mIoU is used as the evaluation index. Best results are marked in bold. GeoKR$^\star$ indicates that it is continuously unsupervised  fine-tuned on the ImageNet pre-training model.}
		\resizebox{\textwidth}{!}{
			\begin{tabular}{c|cccccccccc}
				\toprule
				& 0.25\% & 0.33\% & 0.5\% & 1\% & 2\% & 5\% & 10\% & 20\% & 50\% & 100\% \\
				\midrule
				VGG16 Random &  0.3620 & 0.3026 & 0.3534 & 0.4007 & 0.4240 & 0.4864 & 0.5407 & 0.5814 & 0.6748 & 0.6810  \\ 
				
				ResNet50 Random & 0.3253 & 0.3673 & 0.3495 & 0.4098 & 0.4435 & 0.5340 & 0.5364 & 0.5624 & 0.6522 & 0.6768 \\ 
				
				VGG16 ImageNet & 0.3764 & 0.3303 & 0.4066 & 0.4568 & 0.5132 & 0.5414 & 0.6282 & 0.6175 & 0.6662 & 0.6656 \\
				
				ResNet50 ImageNet & 0.3225 & 0.3864 & 0.3884 & 0.4211 & 0.4637 & 0.5514 & 0.5651 & 0.6088 & 0.6281 & 0.6828 \\
				\midrule
				
				MoCo \cite{he2020momentum} & 0.3570 & 0.3565 & 0.3670 & 0.3984 & 0.4317 & 0.5581 & 0.5986 & 0.6204 & 0.6443 & 0.6683\\
				
				SimCLR \cite{chen2020simple} & 0.3123 & 0.3848 & 0.3905 & 0.3951 & 0.4151 & 0.5527 & 0.5917 & 0.6011 & 0.6362 & 0.6581 \\
				
				BYOL \cite{grill2020bootstrap} & 0.3072 & 0.3568 & 0.3398 & 0.3919 & 0.4222 & 0.5158 & 0.5903 & 0.6004 & 0.6290 & 0.6501 \\
				\midrule
				GeoKR (VGG16) & 0.4010 & 0.4354 & 0.4524 & 0.4924 & 0.5208 & 0.6107 & 0.6433 & 0.6489 & 0.6789 & 0.6951 \\
				GeoKR (ResNet50)  & \textbf{0.4104} & 0.4233 & 0.4528 & 0.4813 & 0.5253 & 0.6226 & 0.6517 & 0.6648 & 0.6848 & 0.6986 \\
				GeoKR$^\star$ (VGG16) & 0.4004 & 0.4251 & 0.4636 & \textbf{0.4975} & 0.5315 & 0.6120 & 0.6532 & 0.6661 & 0.6801 & 0.6953 \\
				
				GeoKR$^\star$ (ResNet50) & 0.4040 & \textbf{0.4388} & \textbf{0.4812} & 0.4955 & \textbf{0.5537} & \textbf{0.6450} & \textbf{0.6704} & \textbf{0.6778} & \textbf{0.6959} & \textbf{0.7048} \\
				\bottomrule
				
		\end{tabular}}
		
		\label{tab:potsdam_segmentation_iou}
	\end{table*}
	
	\subsection{Object Detection}
	
	\textbf{Experimental setup}. We verify the effectiveness of our method in object detection of remote sensing images on the Levir dataset \cite{zou2017random}. It consists of three categories: airplane, ship and oil-tank. We randomly divide the dataset into training set, validation set and testing set according to the ratio of 3:1:1. And we also cut images into slices with the size of $256 \times 256$. We choose mAP (mean average precision) as the evaluation metric.
	We select CenterNet \cite{zhou2019objects} as the evaluation networks. The learning rate is set to 0.005, and will drop to 95\% of the previous every 200 epochs. Meanwhile, after every 200 epochs, we will calculate the accuracy in the validation set, and select the model with the highest accuracy as the final one.  The model is trained for 1000 epochs. We choose mAP (mean average precision) as the evaluation metric. To ensure the object detection method's fundamental performance, we conduct experiments using only ResNet50.
	
	\textbf{Experimental Results}. The detection results are shown in the Table \ref{tab:object_detection_map}. The best results are marked in bold.  Each column represents the results with different proportions of training data. In almost every case, our method yielded the best results, but the improvements are not as noticeable as with scene classification and semantic segmentation. 
	
	All methods have low accuracy when there is a small amount of training data, which indicates that there is a high demand for annotated data in the object detection task. Our method can increase object detection performance over 2\%, but it has a limited impact on the number of annotations. 
	
	\begin{table}[!htb]
		\centering
		\caption{Object detection on Levir dataset. mAP is used as the evaluation index. Best results are marked in bold. GeoKP$^\star$ indicates that it is continuously unsupervised  fine-tuned on the ImageNet pre-training model.}
		\resizebox{\linewidth}{!}{
			\begin{tabular}{c|cccccccccc}
				\toprule
				& 0.5\% & 1\% & 5\% & 10\%  & 50\% & 100\% \\
				\midrule
				
				Random & 0.0192 & 0.0522 & 0.2139 & 0.4678 & 0.7009 & 0.7178 \\
				
				ImageNet  & 0.0175 & 0.0551 & 0.3189 & 0.5250 & 0.7191 & 0.7370\\
				\midrule
				MoCo \cite{he2020momentum}  & 0.0092 & 0.0589 & 0.3425 & 0.5787 & 0.6512 & 0.6826  \\ 
				
				SimCLR \cite{chen2020simple} & 0.0092 & 0.0398 & 0.1632 & 0.5048 &  0.7243 & 0.7229  \\
				BYOL \cite{grill2020bootstrap} & 0.0071 & 0.0452 & 0.2396 & 0.5391 & 0.7215 & 0.7370 \\
				
				\midrule
				
				GeoKR & \textbf{0.0715} & 0.0729 & \textbf{0.3886} & \textbf{0.5979} & 0.7164 & 0.7288  \\
				
				GeoKR$^\star$ & 0.0231 & \textbf{0.0740} &  0.3716 & 0.5969 & \textbf{0.7395} & \textbf{0.7632}  \\
				\bottomrule
				
		\end{tabular}}
		
		\label{tab:object_detection_map}
	\end{table}

	\begin{table*}
		\centering
		\caption{Cloud detection on Levir\_CS dataset. mIoU is used as the evaluation index. Best results are marked in bold. GeoKR$^\star$ indicates that it is continuously unsupervised  fine-tuned on the ImageNet pre-training model.}
		\resizebox{0.9\textwidth}{!}{
			\begin{tabular}{c|cccccccccc}
				\toprule
				& 0.5\% & 1\% & 2\% & 5\% & 10\% & 20\% & 50\% & 100\% \\
				\midrule
				VGG16 Random  & 0.6546 & 0.6402 & 0.6290 & 0.6975 & 0.6705 & 0.6924 & 0.6977 & 0.7148 \\
				ResNet50 Random  & 0.6582 & 0.6364 &  0.6198 & 0.6905 & 0.6764 & 0.6889 & 0.7093 &  0.7320   \\
				
				VGG16 ImageNet  & 0.6683 & 0.6705 & 0.6965 & 0.6726 & 0.7325 & 0.7405 & 0.7351 &  0.7601 
				\\ 
				ResNet50 ImageNet  & 0.6892 & 0.6586 & 0.7077 & 0.6817 & 0.6618 & 0.7219 & 0.7026 & 0.7344  \\
				\midrule
				
				MoCo \cite{he2020momentum}  & 0.6703 & 0.6611 & 0.6920 & 0.6755 & 0.6816 & 0.6853 & 0.7210 & 0.7338   \\ 	
				SimCLR \cite{chen2020simple}  & 0.5994 & 0.6581 & 0.6363 &0.6690  & 0.6532 & 0.7128 & 0.6651 & 0.7256  \\
				BYOL \cite{grill2020bootstrap}  & 0.6712 & 0.6830 & 0.7020 & 0.7078 & 0.6881 & 0.7230 & 0.7349 & 0.7359 \\
				\midrule
				
				GeoKR (VGG16)  & \textbf{0.6992} & 0.6970 & \textbf{0.7176} & 0.7287 & 0.7339 & 0.7348 & 0.7376 & 0.7319
				\\ GeoKR (ResNet50) & 0.6905 & 0.6954 & 0.7130 & 0.7181 & 0.7371 & 0.7563 & 0.7484 & \textbf{0.7622}  \\
				
				GeoKR$^\star$ (VGG16) & 0.6922 & \textbf{0.7088} & 0.7098 & \textbf{0.7440} & \textbf{0.7421} & \textbf{0.7649} & 0.7308 & 0.7537  \\ 
				GeoKR$^\star$ (ResNet50)  & 0.6930 & 0.6989 & 0.7099 & 0.7337 & 0.7233 & 0.7332 & \textbf{0.7510} & 0.7507  \\
				\bottomrule
				
		\end{tabular}}
		
		\label{tab:cloud_detection_iou}
	\end{table*}
	
	\begin{table*}
		\centering
		\caption{Snow detection on Levir\_CS dataset. mIoU is used as the evaluation index. Best results are marked in bold. GeoKR$^\star$ indicates that it is continuously unsupervised fine-tuned on the ImageNet pre-training model.}
		\resizebox{0.9\textwidth}{!}{
			\begin{tabular}{c|cccccccccc}
				\toprule
				
				& 0.5\% & 1\% & 2\% & 5\% & 10\% & 20\% & 50\% & 100\% \\
				\midrule
				VGG16 Random &  0.0502 & 0.2039 & 0.2518 & 0.3308 & 0.3290 & 0.3882 & 0.3996 & 0.4302  \\
				ResNet50 Random &  0.0057 & 0.1578 & 0.2513 & 0.2928 & 0.3190 & 0.4154 & 0.4012 & 0.4447  \\	
				
				VGG16 ImageNet & 0.2147 & 0.2494 & 0.3447 & 0.3273 & 0.4240 & 0.4772 & 0.4662  & 0.5580  \\ 
				ResNet50 ImageNet  & 0.2326 & 0.2743 & 0.3425 & 0.2911 & 0.3115 & 0.3670 & 0.4060 & 0.4700 \\
				\midrule
				
				MoCo \cite{he2020momentum} & 0.2631 & 0.3088 & 0.3176 & 0.3176 & 0.2750 & 0.3288 & 0.4349 &  0.4447 \\ 
				SimCLR \cite{chen2020simple}  & 0.1630 & 0.2599 & 0.2747 & 0.2942 & 0.3237 & 0.2635 & 0.3304 & 0.4031  \\
				
				BYOL \cite{grill2020bootstrap}  & 0.2669 & 0.2704 & 0.3491 & 0.3722 & 0.3558 & 0.4221 & 0.4361 & 0.4486 \\
				\midrule
				
				GeoKR (VGG16) & 0.2496 & 0.3294 & 0.3473 & 0.4230 & \textbf{0.4939} & 0.5194 & 0.5060 & 0.5030  \\
				
				GeoKR (ResNet50) & 0.2025 & 0.3059 & 0.3778 & 0.4070 & 0.4749 & 0.5421 & \textbf{0.5102} &  \textbf{0.5730} \\
				
				GeoKR$^\star$ (VGG16) & 0.2181 & 0.3297 & \textbf{0.4186} & \textbf{0.4902} & 0.4929 & \textbf{0.5485} & 0.4978 & 0.5075  \\ 
				
				GeoKR$^\star$ (ResNet50) & \textbf{0.2532} & \textbf{0.3566} & 0.3938 & 0.4526 & 0.4006 & 0.4628 & 0.5055 & 0.4992  \\
				\bottomrule
				
		\end{tabular}}
		
		\label{tab:snow_detection_iou}
	\end{table*}
	
	\subsection{Cloud / Snow Detection}

	\textbf{Experimental Setup}. The effect of our method on cloud and snow detection is verified by the dataset Levir\_CS \cite{wu2021geographic}. Because cloud and snow detection is a subset of semantic segmentation, we use the same network structure and training strategy as the semantic segmentation section, with the exception that the learning rate is set to 0.001.
	
	\textbf{Experimental Results}. Cloud detection results are shown in Table. \ref{tab:cloud_detection_iou}, while snow detection results are shown in Table. \ref{tab:snow_detection_iou}. The best results are marked in bold.  Each column represents the results with different proportions of training data. 
	
	As a subset of semantic segmentation, there are numerous parallel conclusions between the cloud / snow detection and semantic segmentation experiments. For instance, our methods significantly improved performance, whereas the performance difference between comparative methods is relatively small. However, the experimental results for the cloud detection and snow detection tasks are quite different.
	
	Firstly, cloud detection is generally more accurate than snow detection. The performance improvement for cloud detection is not readily apparent at various dataset scales. Comparing the use of 0.5\% training data and the use of all training data, the performance gap does not exceed 10\%. This also means that our method is restricted in its application to cloud detection, as cloud detection does not require a large amount of annotation data. However, our method can significantly aid in the improvement of snow detection. The improvement can even exceed 10\% in some cases, particularly with less data.

	\subsection{Ablation Experiments}
	
	\begin{table*}
		\centering
		\caption{Ablation studies on RSSCN7 and Vaihingen dataset. Ablations are performed on 1) Classification, 2) Representation, 3) Student and 4) Teacher. }
		\resizebox{\linewidth}{!}{
			\begin{tabular}{cccc|ccccc|ccccc}
				\toprule
				\multicolumn{4}{c}{\textbf{Ablations}} &
				\multicolumn{5}{c}{\textbf{RSSCN7}}  &  \multicolumn{5}{c}{\textbf{Vaihingen}}  \\
				Classification & Representation & Student & Teacher & 5\% & 10\% & 20\% & 50\% & 100\% & 0.33\% & 1\% & 5\% & 20\% & 100\% \\
				\midrule
				$\times$ & $\times$ & $\times$ & $\times$ & 0.6752 & 0.7829 & 0.8276 & 0.8593 & 0.9229 & 0.3424 & 0.3470 & 0.4640 & 0.5177 & 0.7015\\
				$\checkmark$ & $\times$ & $\times$ & $\times$ & 0.7976 & 0.8395 & 0.8833 & 0.9133 & 0.9281 & 0.2942 & 0.4756 & 0.6249 & 0.6976 & 0.7236\\
				$\times$ & $\checkmark$ & $\times$ & $\times$ & 0.8071 & 0.8500 & 0.8881 & 0.8943 & 0.9376 & 0.2926 &  0.4777 & 0.6150 & \textbf{0.7104} & 0.7322\\
				$\times$ & $\checkmark$ & $\checkmark$ & $\times$  & 0.8310 & 0.8790 & \textbf{0.8971} & \textbf{0.9305} & 0.9357 & 0.3178 &  0.4992 & 0.6377 & 0.7038 & \textbf{0.7374}\\
				$\times$ & $\checkmark$ & $\times$ & $\checkmark$ & \textbf{0.8448} & \textbf{0.8933} & 0.8933 & 0.9152 & \textbf{0.9419} & \textbf{0.4138} & \textbf{0.5150} & \textbf{0.6390} & 0.6978 & 0.7268\\
				\bottomrule
		\end{tabular}}
		\label{tab:ablation_experiments}
	\end{table*}
	
	We design the ablation experiment to determine the effect of the proposed method's various components. Table \ref{tab:ablation_experiments} summarizes the experimental findings. We validate the effectiveness of pre-training on the RSSCN7 and Vaihingen datasets, respectively, using ResNet50 as the backbone. The baseline model is based on ResNet50 and fine-tuned using the ImageNet pre-training model. Each ablation item is described in detail below:
	
	\begin{itemize}
		\item \textbf{Classification}. Pre-training is accomplished through the use of classification method. To begin, the land cover type with the greatest proportion in the image is chosen as the category label, and an ImageNet-like classification dataset is constructed. Then, a classification layer is added to the ResNet50 backbone to complete the network pre-training. It can also be regarded as the simplest method to generate knowledge representations.
		\item \textbf{Representation}. The knowledge representation is used to provide supervision information, and the KL loss is used to guide  the network pre-training. The ResNet50 backbone and projection layer continue to be used.  Teacher networks is not adopted. 
		\item \textbf{Student}. The teacher networks is added to further constrain the training of student networks, but the student model is used as the pre-training model on downstream tasks.
		\item \textbf{Teacher}. The teacher networks is added to further constrain the training of student networks, and the teacher model is also used as the pre-training model on downstream tasks.
	\end{itemize}
	
	It can be seen that with the geographical knowledge, even the simplest methods for generating knowledge representation can significantly improve the effectiveness of the pre-training model. It fully demonstrates the necessity and effectiveness of geographical knowledge for representation learning in remote sensing images. But the method we adopt to generate knowledge representations in this paper and teacher networks can further improve the performance of pre-training. The best results are achieved with teacher model. Moreover, it is of little effect whether to select student model or teacher model, but both are better than that without the constrain of teacher networks. It is proved that the teacher networks can really play a guiding role during training, and remove the impact on differences of imaging time between remote sensing images and the geographical knowledge. 
	
	\section{Conclusion}\label{section:conclusion}
	We propose a geographical knowledge-driven representation learning method for remote sensing images (GeoKR). We employ geographical knowledge (including global land cover products and the geographical location of remote sensing images) to supervise representation learning and network pre-training using millions of unlabeled remote sensing images. Due to the difference in imaging times and resolutions between remote sensing images and geographical knowledge, generated supervision information may contain noises. We develop an efficient pre-training framework based on mean-teacher networks and complete network pre-training by aligning the network's image representations with the knowledge representations generated using geographical knowledge. In addiction,  Levir-KR dataset is build up with 1,431,950 images from Gaofen satellites. On downstream tasks such scene classification, semantic segmentation, object detection, and cloud / snow detection, our proposed method outperforms ImageNet pre-training and self-supervised representation learning and can effectively reduce the burden of data annotation. Our method can improve semantic segmentation and object detection accuracy by more than 2\% when using all training data. When training data is limited, improvements are even more. To achieve the effect of ImageNet pre-training with total training data, our method only requires approximately half of the training data in the scene classification task and approximately 20\% of the training data in the semantic segmentation task. All experiments confirm that our proposed method can be used as a novel paradigm for networks pre-training.

	{\small
		\bibliographystyle{IEEEtran}
		\bibliography{main}
	}

	%

	
	\vfill
	


\end{document}